\pdfoutput=1
\documentclass[10pt,twocolumn,letterpaper]{article}

\usepackage{cvpr}
\usepackage{graphicx}
\DeclareGraphicsExtensions{.ps,.pdf,.png,.tif,.jpg}

\usepackage{times}
\usepackage{epsfig}
\usepackage{graphicx}
\usepackage{amsmath}
\usepackage{amssymb}

\usepackage{times}

\usepackage{color}
\usepackage{amsmath}
\usepackage[ruled,vlined]{algorithm2e}
\usepackage{siunitx}
\usepackage{authblk}
\usepackage{multirow}

\DeclareMathOperator*{\argminC}{\arg\min} 

\cvprfinalcopy 

\def\cvprPaperID{10} 

\begin{document}

\title{Illumination-based Transformations Improve Skin Lesion Segmentation in Dermoscopic Images}


\author[]{Kumar Abhishek\thanks{Corresponding author} }
\author[]{Ghassan Hamarneh}
\author[]{Mark S. Drew}
\affil[]{School of Computing Science, Simon Fraser University, Canada\protect \\\texttt{\{kabhishe, hamarneh\}@sfu.ca, mark@cs.sfu.ca}}

\maketitle

\begin{abstract}
   The semantic segmentation of skin lesions is an important and common initial task in the computer aided diagnosis of dermoscopic images. Although deep learning-based approaches have considerably improved the segmentation accuracy, there is still room for improvement by addressing the major challenges, such as variations in lesion shape, size, color and varying levels of contrast. In this work, we propose the first deep semantic segmentation framework for dermoscopic images which incorporates, along with the original RGB images, 
   information extracted using the physics of skin illumination and imaging.
   In particular, we incorporate information from specific color bands, illumination invariant grayscale images, and shading-attenuated images. We evaluate our method on three datasets: the ISBI ISIC 2017 Skin Lesion Segmentation Challenge dataset, the DermoFit Image Library, and the PH2 dataset and observe improvements of 12.02\%, 4.30\%, and 8.86\% respectively in the mean Jaccard index over a baseline model trained only with RGB images.
\end{abstract}

\section{Introduction}

Skin conditions are the most common reason for visits to general practitioners in studied populations~\cite{schofield2011skin}, and the prevalence of skin cancer in the United States has been higher than all other cancers combined over the last three decades~\cite{rogers2015incidence}. Melanoma, a type of skin cancer which represents only a small fraction of all skin cancers in the USA, is responsible for over 75\% of skin cancer related fatalities and over 10,000 deaths annually in the USA alone~\cite{SkinCancer}. However, studies have shown that early diagnosis can drastically improve patient survival rates. While skin cancers can be diagnosed by visual examination, it is often difficult to distinguish malignant lesions from healthy skin. As a result, computer aided diagnoses of skin lesions have been widely used to automate the assessment of dermoscopic images. The segmentation of skin lesion images is therefore a crucial step in the diagnosis and subsequent treatment. Segmentation refers to the process of delineating the lesion boundary by assigning pixel-wise labels to the dermoscopic images, so as to separate the lesion from the surrounding healthy skin. However, this is a complicated task, primarily because of the large variety in the shape, color, presentation, and contrast of skin lesions, originating from intra- and inter-class variations as well as image acquisition.


Recent years have witnessed the successful applications of machine learning, particularly deep learning-based approaches, to the semantic segmentation of skin lesions. Numerous contributions have been made in terms of new architectures (such as fully convolutional network models~\cite{Yuan}, deep residual networks~\cite{yu2017automated}, deep auto-context architectures~\cite{mirikharaji2018deep}, etc.), shape~\cite{mirikharaji2018star} and texture~\cite{zhang2019automatic} priors, input transformations~\cite{taghanaki2019improved}, synthesis-based augmentations~\cite{Pollastri2019, abhishek2019mask2lesion}, and loss functions~\cite{abraham2019novel}.


Although deep learning-based approaches have made significant improvements to the segmentation performance, they are reliant on a large amount of training data in order to yield acceptable performances. Deep learning-based approaches also tend to ignore knowledge about illumination in skin lesion images and other such physics-based properties, an area that has been explored in the past. Madooei et al.~\cite{madooei2012intrinsic} proposed a new 2D log-chromaticity color space and showed that color intensity triplets in skin images lie on a plane, and used Otsu's algorithm to segment skin lesions, demonstrating superior performance even on low-contrast lesions. In another work~\cite{madooei2012automated}, they also presented pre-processing techniques for improved segmentation of skin lesions. They calculated an illumination invariant grayscale `intrinsic' image and used it to attenuate shading and lighting intensity changes in dermoscopic images. Moreover, they also presented a novel RGB-to-grayscale conversion algorithm for dermoscopic images using principal component analysis in the optical density space. In a more recent work, Guarracino et al.~\cite{guarracino2018sdi+} proposed an unsupervised approach for skin lesion segmentation from dermoscopic images by choosing certain color bands. Ng et al.~\cite{hua2019effect} demonstrated an improvement in segmentation performance with the use of color constancy algorithms in a fully convolutional network-based segmentation model. However, very little research has been done on the applicability of color image theory and illumination information to a deep learning-based semantic segmentation framework.

We propose a novel deep semantic segmentation algorithm for dermoscopic images, 
which leverages prior illumination and color theory knowledge. In particular, we 
build upon previous works and leverage specific color bands, intrinsic images, and skin image information to yield improved segmentation results. To the best of our knowledge, this is the first work that incorporates such information in a deep learning-based framework.

The rest of the paper is structured as follows: Section~\ref{sec:Method} describes the proposed approach, Section~\ref{sec:Experiments} describes the dataset, the experiments, and the evaluation metrics, Section~\ref{sec:Results} contains quantitative and qualitative analyses of the proposed approach, and Section~\ref{sec:Conclusion} concludes the paper.

\section{Method}    \label{sec:Method}

\begin{figure}[!h]
\centering
\includegraphics[width=0.45\textwidth]{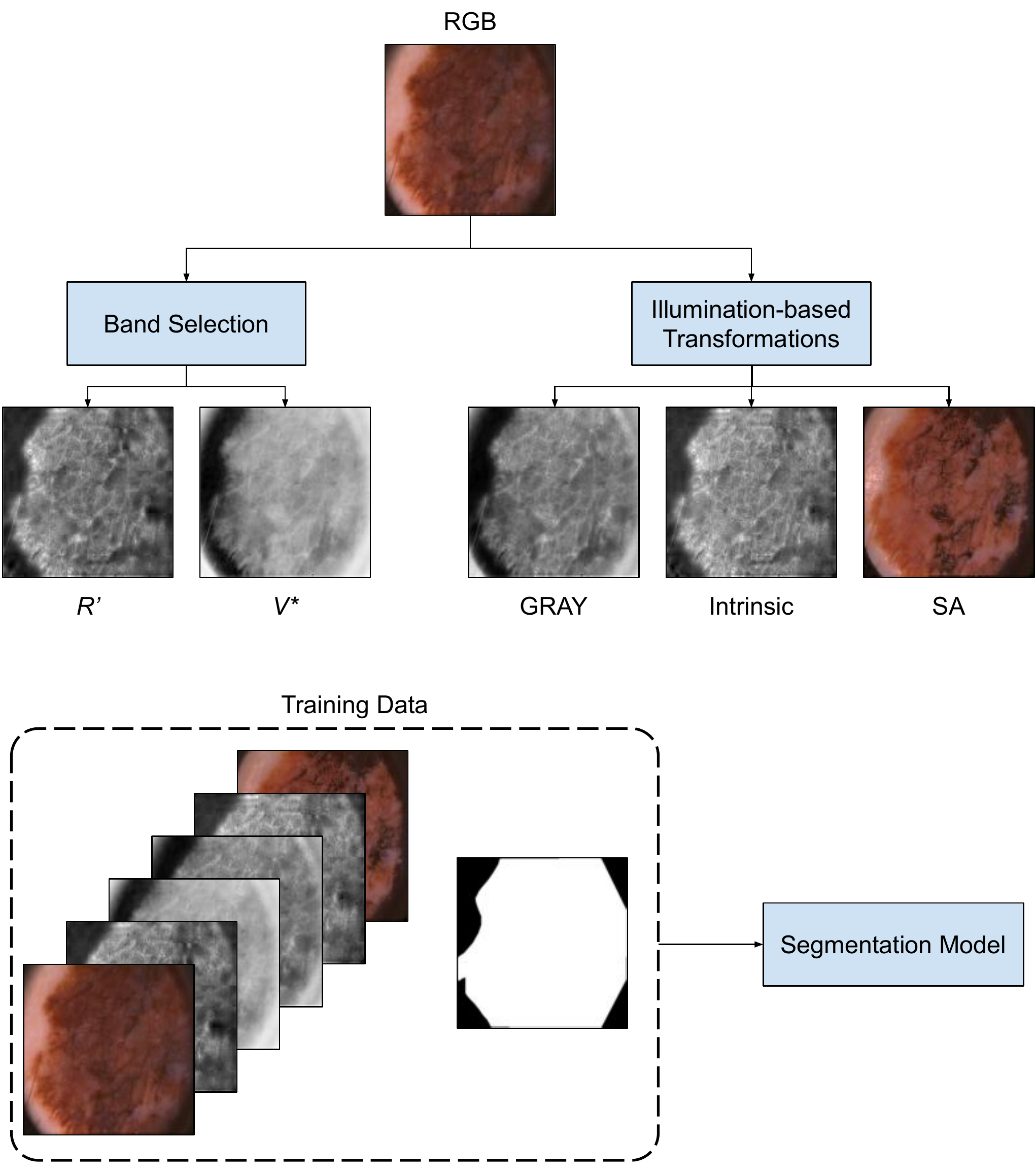}
\caption{An overview of the proposed approach. Various color bands and transformations are computed as explained in Section~\ref{sec:Method} and concatenated channel-wise to the original RGB dermoscopic image in order to train the segmentation model.}
\label{fig:overview}
\end{figure}

In this work, we extract color information from the RGB dermoscopic images and use them along with the original image to train a deep semantic segmentation model. In particular, we use (a) variations of certain color bands (Section~\ref{subsec:BANDS}), (b) a color-theory-based grayscale estimate (Section~\ref{subsec:GRAY}), (c) an illumination-invariant intrinsic grayscale image (Section~\ref{subsec:INTRINSIC}), and (d) a shading-attenuated image obtained from the dermoscopic image (Section~\ref{subsec:SHADING}). Figure~\ref{fig:overview} shows an overview of the proposed approach In the subsequent sections, we describe the methods for obtaining these images.

\subsection{Choosing color bands}   \label{subsec:BANDS}
We choose the two color bands which have been shown to be efficient for skin lesion segmentation~\cite{guarracino2018sdi+}: 
the red color channel from the normalized RGB image, and the complement of the value channel from the HSV color space representation of the image (denoted by $R'$ and $V^*$ respectively) and concatenate them to the original RGB dermoscopic image. They are defined as:
\begin{equation}
    R' = \frac{R}{R + G + B}, \ \textrm{and}
\end{equation}
\begin{equation}
    V^* = 1 - V,
\end{equation}

\noindent where $R, G, B$ denote the channels from the original image and $V$ denotes the Value channel from the HSV representation of the image. For computational efficiency, instead of converting the image from the RGB to the HSV color space, the $V$ channel can directly be calculated as:
\begin{equation}
    V = max\ (\frac{R}{M}, \frac{G}{M}, \frac{B}{M}),
\end{equation}

\noindent where $M = 2^n - 1$ denotes the number of gray-levels in an $n$-bit image ($M=255$ for our 8-bit color images).

\subsection{Intrinsic images}   \label{subsec:INTRINSIC}
We follow the approach proposed by Finlayson et al.~\cite{finlayson2004intrinsic} to derive an illumination-invariant grayscale `intrinsic' image using entropy minimization. Given an RGB image, let $R_{k}, k \in \{1,2,3\}$ denote the channel-wise intensities, and the 3-vector chromaticities can be obtained by dividing each color channel by the geometric mean of the channels.
\begin{equation}    \label{eqn:chromaticity}
    c_k = \frac{R_k}{R_p} \ \ \forall k \in \{1,2,3\},
\end{equation}

\noindent where $R_p = \sqrt[\leftroot{-2}\uproot{2}3]{\prod_{k=1}^{3} R_k}$. Finlayson et al. note that while it is possible to obtain 2-vector chromaticities by dividing by one of the color channels, the choice of dividing by the geometric mean ensures that there is no bias towards any particular channel.

From~\cite{finlayson2004intrinsic}, assuming the light as a Planckian radiator (and using Wein's approximation~\cite{wyszecki2000color}) and the camera sensors to be fairly narrow-band, the channel $R_k$ can be written as:
\begin{equation}    \label{eqn:R_exp}
    R_k = \sigma \ L \ k_1 \ \lambda_k^{-5} \exp\left({-\frac{k_2}{T\lambda_k}}\right) S(\lambda_k) q_k,
\end{equation}

\noindent 
where $\sigma$ is the Lambertian shading, $L$ is the overall light intensity, $T$ is the temperature of the lighting color, $S(\lambda_k)$ is the spectral reflectance of the surface (which is the skin in our case) as a function of the wavelength $\lambda_k$, $q_k$s are the camera sensor sensitivity functions, and $k_1, k_2$ are constants. 
Therefore, the log-chromaticities (obtained by taking the logarithm of Eqn.~\ref{eqn:chromaticity}) can be written as:
\begin{equation}    \label{eqn:log-chromaticity}
    \rho_k \equiv \textrm{log}(c_k) = \textrm{log} \left(\frac{s_k}{s_p}\right) + \left(\frac{e_k - e_p}{T}\right),
\end{equation}

\noindent where $s_k = k_1 \lambda_k^{-5} S(\lambda_k) q_k$ and $e_k = -k_2/\lambda_k$. Note that this expression does not have the shading and the intensity information. Eqn.~\ref{eqn:log-chromaticity} is the parametric equation of a straight line with $T$ as the parameter, and although the surface information is present in the intercept of the line, the direction is given by $e \equiv (e_k - e_p)$, which is independent of the surface.

With 2D log-chromaticities, it is possible to obtain the intrinsic image by projecting $\rho_k$ in a direction orthogonal to $e$ (denoted by $e^\perp$), followed by taking its exponential. 
However, dividing by the geometric mean (Eqn.~\ref{eqn:chromaticity}) yields 3D log-chromaticities, and therefore, the task is to find a projector $P$ which can project $\rho$ onto the 2D chromaticity space, which is a plane. Note that the log-chromaticities $\rho$ are orthogonal to $u \equiv (1,1,1)^{T} / \sqrt{3}$, and so the projector $P^\perp_u$ can be used to characterize the plane.
Since $P^\perp_u$ has two non-zero eigenvalues, it can be decomposed as:

\begin{equation}
    P^\perp_u = I - uu^T = U^TU,
\end{equation}

\noindent where $I$ is the identity matrix and $U$ is a $2\times3$ orthogonal matrix, which projects the three $\rho$ vectors onto a coordinate system in the plane as two vectors denoted by $\chi$. It should be noted that straight lines in $\rho$ remain straight in $\chi$.

\begin{equation}
    \chi \equiv U \rho
\end{equation}


The next step is to find the optimal angle $\theta_{\textrm{proj}}$ to project along in the $\{\chi_1, \chi_2\}$ plane, for which the entropy for the marginal distribution along a 1D line orthogonal to the lighting direction is minimized. The resulting projected log grayscale image $\mathcal{I}_{\textrm{projected}}$ is given by

\begin{equation}
    \mathcal{I}_{\textrm{projected}} = \chi_1 \ \textrm{cos}  \theta_{\textrm{proj}} + \chi_2 \ \textrm{sin}  \theta_{\textrm{proj}}
\end{equation}

To compute the best projection angle, only the middle 90\% of the data is used. This is done to exclude the outliers by using data between the $5^{\mathrm{th}}$ and the $95^{\mathrm{th}}$ percentiles. Then, Scott's rule~\cite{scott2012multivariate} is used to estimate the bin width for constructing the histogram as:

\begin{equation}
    \textrm{bin width} = 3.5 * \textrm{STD (projected data)} * \sqrt[\leftroot{-2}\uproot{2}3]{N},
\end{equation}

\noindent where STD($\cdot$) denotes the standard deviation and $N$ is the size of the grayscale image data for a given angle $\omega$. Next, for each angle, probabilities $p_i$ for each bin $i$ are computed by dividing the bin by the sum of the bin counts, and the entropy is calculated as:

\begin{equation}
    \eta = - \sum_i p_i(I) \textrm{log}(p_i(I))
\end{equation}

The angle which yields the lowest entropy is chosen as the projection angle, and finally the projected log-image $\mathcal{I}_{\textrm{projected}}$ is exponentiated to yield the intrinsic image. The entire approach is shown in Algorithm~\ref{alg:entropy}.

\begin{algorithm}[]
\SetAlgoLined
\KwIn{RGB image $\mathcal{I}$}
\KwOut{Grayscale intrinsic image $I_{\textrm{Intrinsic}}$}
\medskip
 construct 2D log-chromaticity representation of $\mathcal{I}_{\textrm{projected}}$
 \For{$\omega \gets 1$ \textbf{to} $180$} {
  calculate histogram bin width\;
  compute histogram with middle 90\% data\;
  compute $\eta_\omega$, the entropy for the angle $\omega$\;
}
$\theta_{\textrm{proj}} \gets \argminC_\omega \eta_\omega$\;
$\mathcal{I}_{\textrm{projected}} \gets \chi_1 \ \textrm{cos}  \theta_{\textrm{proj}} + \chi_2 \ \textrm{sin}  \theta_{\textrm{proj}}$\;
$I_{\textrm{Intrinsic}} \gets \exp(\mathcal{I}_{\textrm{projected}})$

\Return{$I_{\textrm{Intrinsic}}$}\;
 \caption{Intrinsic image by entropy minimization}
 \label{alg:entropy}
\end{algorithm}

\subsection{Grayscale images of skin lesions}   \label{subsec:GRAY}

Madooei et al.~\cite{madooei2012automated} proposed a RGB-to-grayscale conversion algorithm for dermoscopic images based on the optics and the reflectance properties of skin surfaces. Unlike a traditional grayscale representation calculated as the weighted sum of the red, the green, and the blue channels~\cite{poynton1997frequently}, this grayscale image preserves the lesion while suppressing the healthy skin, thereby increasing the contrast between the healthy and the affected regions.
Based on the skin models proposed by Hiraoka et al.~\cite{hiraoka1993monte} and Tsumura et al.~\cite{tsumura1999independent}, the spectral reflection of skin at a pixel $(x,y)$ under polarized light can be written as:
\begin{equation}
    \begin{split}
        S(x,y,\lambda) = \exp\Big[
        & -\rho_m(x,y)\alpha_m(\lambda)l_m(\lambda)  \\
        & -\rho_h(x,y)\alpha_h(\lambda)l_h(\lambda) \Big],
\end{split}
\end{equation}

\noindent where $\rho_{\{m,h\}}, \alpha_{\{m,h\}}, \textrm{and } l_{\{m,h\}}$ denote the densities ($\textrm{cm}^{-3}$), cross-sectional areas ($\textrm{cm}^{2}$) for scattering absorption, and mean path lengths for photons in the epidermis and dermis layers of the human skin for melanin and hemoglobin (denoted by subscripts $m$ and $h$ respectively). Substituting this expression in Eqn.~\ref{eqn:R_exp} followed by taking the logarithms on both sides yields:
\begin{equation}    \label{eqn:skin_2D}
\begin{split}
    \textrm{log}(R_k(x,y)) &= - \rho_m(x,y) \sigma_m - \rho_h(x,y) \sigma_h \\
    & + \textrm{log}\left(k_1 I(x,y) \sigma(x,y)\right) \\
    & + \textrm{log}\left(\lambda_k^{-5}\right) - \frac{k_2}{\lambda_k T},
\end{split}
\end{equation}

\noindent where $\sigma_{\{m,h\}} = \alpha_{\{m,h\}}l_{\{m,h\}}$. 
Eqn.~\ref{eqn:skin_2D} suggests that the pixels from a skin image lie on a plane in the optical density space described by $\left[ - \textrm{log}R_1, - \textrm{log}R_2, - \textrm{log}R_3 \right]$. As such, Madooei et al. observe that in almost all the skin lesion images analyzed, the first eigenvector in the principal component analysis (PCA) explains a very high fraction of the total variance, and thus contains most of the information in the image. As such, the first principal component can be used to obtain a grayscale skin lesion image. The approach has been described in Algorithm~\ref{alg:gray}.

\begin{algorithm}[]
\SetAlgoLined
\KwIn{RGB image $\mathcal{I}_{\mathrm{RGB}}$}
\KwOut{Grayscale image $\mathcal{I}_{\mathrm{GRAY}}$}
\medskip
 
 $\mathcal{I'}_{\mathrm{RGB}} \gets$ channel-wise-flatten$(\mathcal{I}_{\mathrm{RGB}})$\;
 $\mathcal{I}^*_{\mathrm{RGB}} \gets \mathcal{I'}_{\mathrm{RGB}}$ image in optical density space\;
 $P \gets \textrm{1st principal component from PCA} (\mathcal{I}^*_{\mathrm{RGB}})$\;
 $\mathcal{I}_{\mathrm{GRAY}} \gets$ reshape($P$)
 
 \Return{$\mathcal{I}_{\mathrm{GRAY}}$}\;
 \caption{Grayscale skin lesion image}
 \label{alg:gray}
\end{algorithm}

\subsection{Shading-attenuated skin lesion images}   \label{subsec:SHADING}

The non-flat nature of skin surfaces, especially lesions, and the effect of light intensity falloffs towards the edges of the skin lesions can induce shading in dermoscopic images, which can degrade the segmentation (and classification) performance. Madooei et al.~\cite{madooei2012automated} proposed to use the intrinsic images generated by Finlayson et al.~\cite{finlayson2004intrinsic} to perform illumination normalization in dermoscopic images, thereby performing shading-attenuation. They proposed to use the intrinsic image to normalize the intensity values. Given a dermoscopic image, its intrinsic image is first calculated. The RGB image is then converted to the HSV color space. In order to normalize the intensities, the Value (V) channel of the HSV image is used. First, both the intrinsic image and the Value channel are normalized. The image intensity histogram of the Value channel is then mapped to that of the intrinsic image. Finally, this new normalized and histogram-mapped Value channel is used to replace the original Value channel in the HSV image, and the resultant image is then mapped back to the RGB color space. The authors demonstrated a significant attenuation in the shading and the intensity falloff using this approach. The entire approach is summarized in Algorithm~\ref{alg:shadow}.

\begin{algorithm}[]
\SetAlgoLined
\KwIn{RGB image $\mathcal{I}_{\mathrm{RGB}}$}
\KwOut{Shading-attenuated RGB image $\mathcal{I}_{\mathrm{SA}}$}
\medskip
 
 compute $I_{\textrm{Intrinsic}}$ from $\mathcal{I}_{\mathrm{RGB}}$ using Algorithm~\ref{alg:entropy}\;
 $\mathcal{I}_{\textrm{HSV}} \gets$ \texttt{rgb2hsv}($\mathcal{I}_{\mathrm{RGB}}$)\;
 normalize $I_{\textrm{Intrinsic}}$, $V$\;
 $V^{*} \gets $ $V$ histogram-matched to $I_{\textrm{Intrinsic}}$\;
 $\mathcal{I}_{\mathrm{SA}} \gets$ \texttt{hsv2rgb}($\mathcal{I}_{\mathrm{HSV^{*}}}$)\\
 
 \Return{$\mathcal{I}_{\mathrm{SA}}$}\;
 \caption{Shading-attenuated skin lesion image}
 \label{alg:shadow}
\end{algorithm}

\begin{figure*}[!h]
\centering
\includegraphics[width=0.9\linewidth]{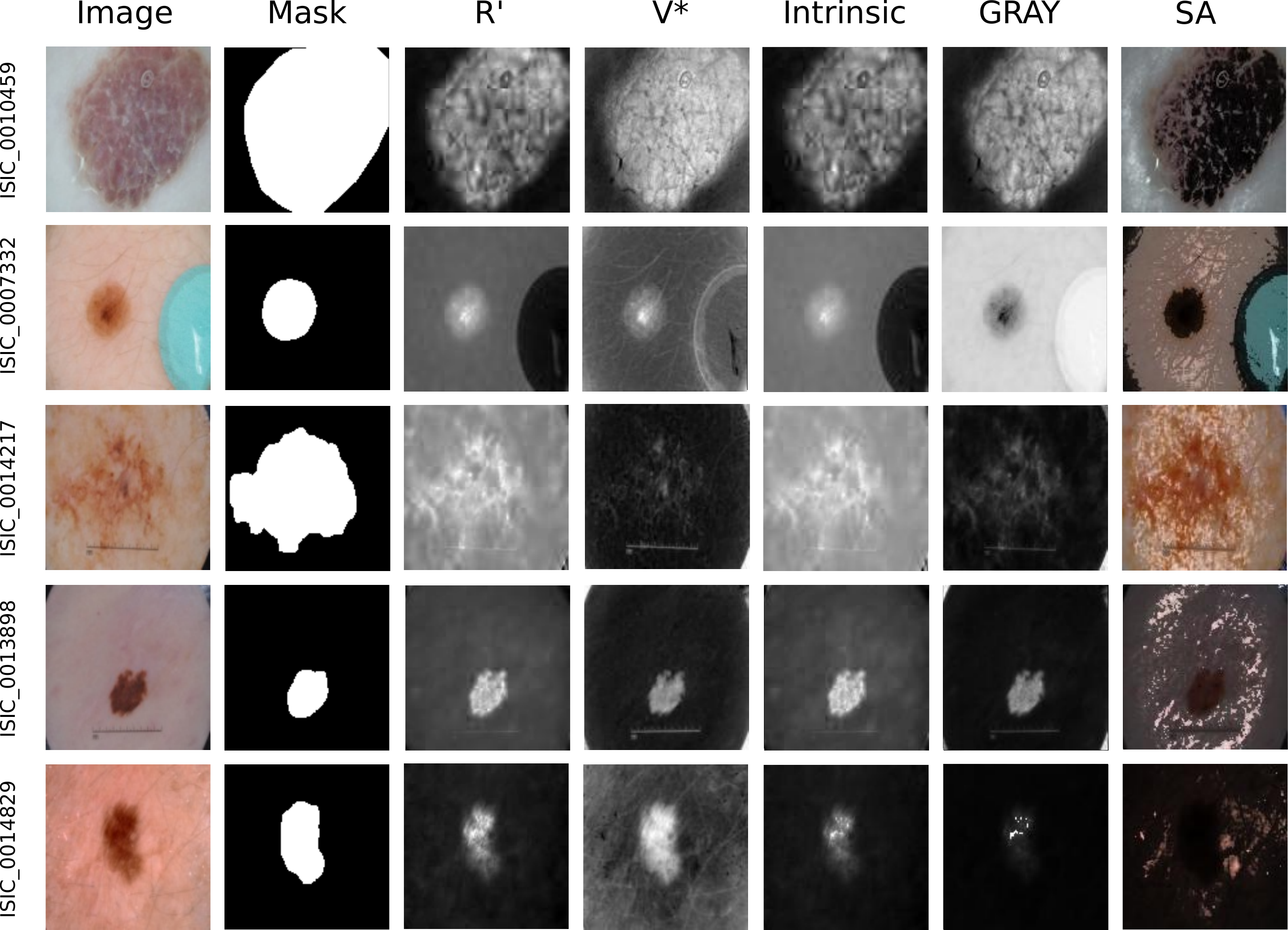}
\caption{Transformation results for 5 images from the ISIC 2017 training set.}
\label{fig:TransformedImagesGrid}
\end{figure*}

\section{Datasets and Experimental Details}    \label{sec:Experiments}

\subsection{Datasets}

We evaluate our proposed approach on three skin lesion image datasets, namely the ISIC ISBI 2017 dataset, the DermoFit dataset, and the PH2 dataset.


\subsubsection{ISBI ISIC 2017}

The ISIC 
ISBI 
2017: Skin Lesion Analysis Towards Melanoma Detection Challenge~\cite{codella2018skin} Part 1 dataset contains
skin lesion images and the corresponding manually annotated lesion delineations belonging to three diagnoses: melanoma, seborrheic keratosis, and benign nevi. The dataset is split into training, validation, and testing subsets containing 2000, 150, and 600 images respectively.

\subsubsection{DermoFit}

The DermoFit Image Library~\cite{ballerini2013color,dermofit2013website} contains 1300 skin lesion images belonging to ten diagnosis classes, along with the corresponding binary segmentation masks. We divide the original dataset into training, validation, and testing splits in the ratio of $60:10:30$.

\subsubsection{PH2}

The PH2 Image Database~\cite{mendoncca2013ph} contains a total of 200 dermoscopic images of common nevi, atypical nevi, and melanomas, along with their lesion segmentations annotated by an expert dermatologist.

\begin{table*}[!h]
\centering
\caption{Quantitative results for the seven methods on 600 images from the ISIC 2017 test set. (mean $\pm$ standard error).}
\label{tab:results}
{\renewcommand{\arraystretch}{1.5}
\begin{tabular}{cccccc}
\hline
\textbf{Method}         & \textbf{Accuracy}        & \textbf{Dice Coefficient} & \textbf{Jaccard Index}   & \textbf{Sensitivity} & \textbf{Specificity} \\ \hline
RGB Only & $0.9029 \pm 0.0053$ & $0.7781 \pm 0.0086$ & $0.6758 \pm 0.0095$ & $0.7471 \pm 0.0091$ & $0.9683 \pm 0.0031$\\
All Channels & $\textbf{0.9220} \pm \textbf{0.0045}$ & $\textbf{0.8386} \pm \textbf{0.0078}$ & $\textbf{0.7570} \pm \textbf{0.0089}$ & $\textbf{0.8706} \pm \textbf{0.0077}$ & $0.9516 \pm 0.0037$\\
No $R'$ & $0.9185 \pm 0.0046$ & $0.8243 \pm 0.0078$ & $0.7363 \pm 0.0090$ & $0.7949 \pm 0.0085$ & $0.9735 \pm 0.0030$\\
No $V^*$ & $0.9189 \pm 0.0049$ & $0.8263 \pm 0.0077$ & $0.7381 \pm 0.0089$ & $0.7892 \pm 0.0087$ & $0.9786 \pm 0.0025$\\
No Intrinsic & $0.9092 \pm 0.0056$ & $0.7997 \pm 0.0094$ & $0.7139 \pm 0.0103$ & $0.7662 \pm 0.0104$ & $\textbf{0.9803} \pm \textbf{0.0024}$\\
No GRAY & $0.9116 \pm 0.0052$ & $0.8163 \pm 0.0080$ & $0.7260 \pm 0.0091$ & $0.8041 \pm 0.0090$ & $0.9643 \pm 0.0033$\\
No SA & $0.9198 \pm 0.0050$ & $0.8274 \pm 0.0083$ & $0.7445 \pm 0.0093$ & $0.8137 \pm 0.0088$ & $0.9603 \pm 0.0044$\\
\hline
\end{tabular}%
}
\end{table*}

\begin{figure*}[!h]
\centering
\includegraphics[width=0.9\textwidth]{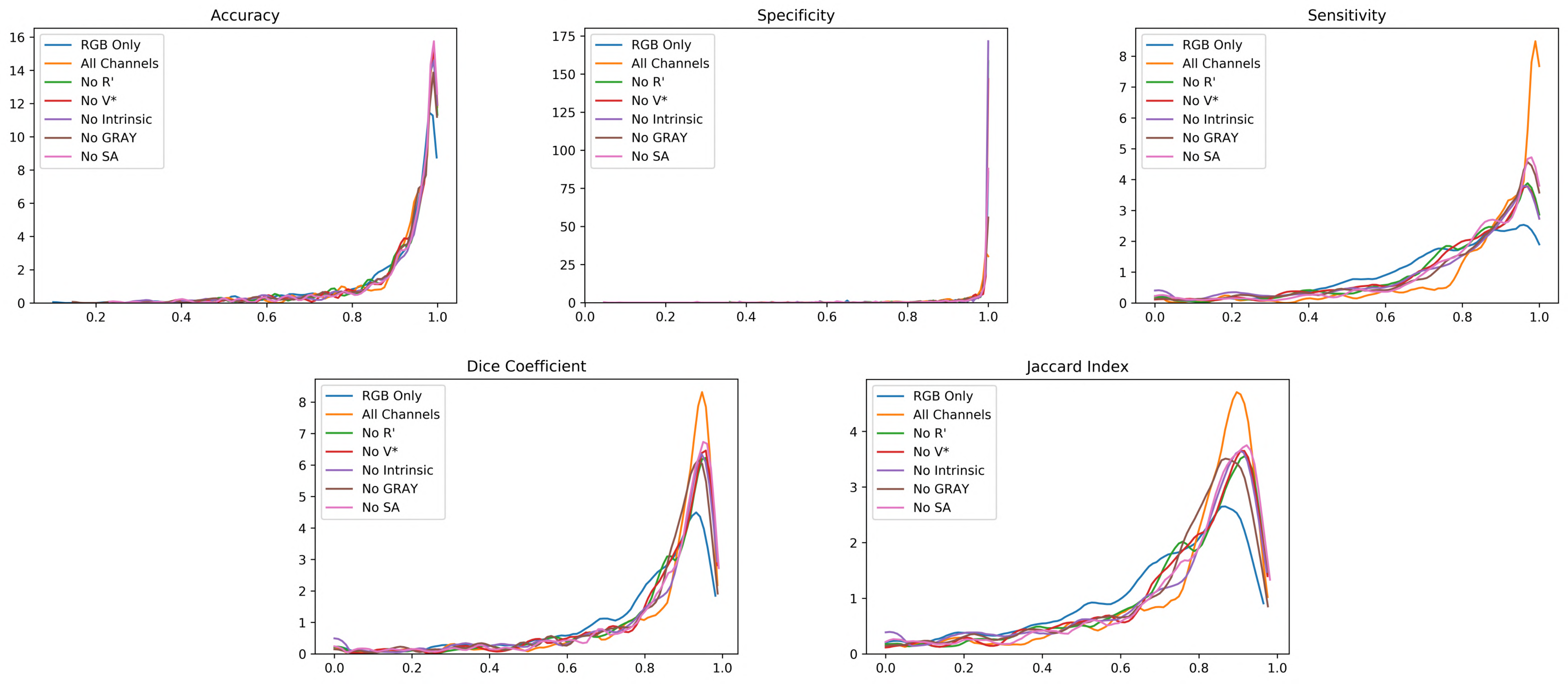}
\caption{Kernel density estimates for the five metrics for all the segmentation methods evaluated on the ISIC 2017 test set.}
\label{fig:plots}
\end{figure*}

\subsection{Experiments and Evaluation}

Since the goal of this work is to demonstrate the effectiveness of the various color theory and illumination-based transformations for enhancing segmentation performance, we use U-Net~\cite{ronneberger2015u} as the baseline segmentation network. The U-Net consists of a symmetric encoder-decoder architecture, with skip connections between symmetrically corresponding layers in the encoder and the decoder, which help in recovering the full spatial resolution~\cite{li2018visualizing} and address the problem of gradient vanishing~\cite{taghanaki2019deep}. For evaluating upon the ISIC 2017 dataset, we train seven segmentation models where the inputs to the corresponding networks are the following:
\begin{itemize}
  \itemsep 0pt
    \item \textbf{RGB Only}: The original 3-channel RGB dermoscopic image.
    \item \textbf{All Channels}: The original RGB dermoscopic image channel-wise concatenated with $R', V^*$ (Section~\ref{subsec:BANDS}), intrinsic image (denoted by Intrinsic; Section~\ref{subsec:INTRINSIC}), grayscale image (denoted by GRAY; Section~\ref{subsec:GRAY}), and shading-attenuated image (denoted by GRAY; Section~\ref{subsec:SHADING}). The result is a 10-channel image.
    \item Next, to determine the contribution of each of the transformations described in Section~\ref{sec:Method}, we drop one component at a time from the 10-channel image above, and denoted it by \textbf{No \underline{x}}, where \textbf{x} denotes the dropped channel. The models are:
    \begin{itemize}
        \item \textbf{No $R'$}: 9-channel image.
        \item \textbf{No $V^*$}: 9-channel image.
        \item \textbf{No GRAY}: 9-channel image.
        \item \textbf{No Intrinsic}: 9-channel image.
        \item \textbf{No SA}: 7-channel image.
    \end{itemize}
\end{itemize}


For each of these aforementioned models, the input layer of the segmentation network is modified to handle the corresponding number of channels, and the rest of the architecture remains the same. The models are trained to predict the pixelwise labels for the semantic segmentation task. All images and their corresponding ground truth segmentation masks are resized to $128\times128$ using nearest neighbor interpolation from Python's SciPy library. All networks are trained with Dice loss~\cite{milletari2016v} using mini-batch stochastic gradient descent with a batch size of $40$ (since a larger batch size exceeded our GPU memory) and a learning rate of $1e-3$. We apply real time data augmentation strategies (random horizontal and vertical flips and rotations) during training. All the code was written in Python and the PyTorch framework was used to implement the deep segmentation models.

For the evaluation of the three methods, we report the metrics used by the official challenge - pixel-wise accuracy, sensitivity, specificity, Dice similarity coefficient, and Jaccard index (also known as the intersection over union). They are given by:

\begin{equation}
    \textrm{Accuracy} = \frac{TP+TN}{TP + TN + FP + FN},
\end{equation}
\begin{equation}
    \textrm{Sensitivity} = \frac{TP}{TP + FN},
\end{equation}
\begin{equation}
    \textrm{Sensitivity} = \frac{TN}{TN + FP},
\end{equation}
\begin{equation}
    \textrm{Dice coefficient}, Dice(A,B) = 2\ \frac{\left|A \cap B\right|}{\left|A\right| + \left|B\right|},
\end{equation}
\begin{equation}
    \textrm{Jaccard index}, J(A,B) = \frac{\left|A \cap B\right|}{\left|A \cup B\right|},
\end{equation}

where \textit{TP, TN, FP, FN} denote true positive, true negative, false positive, and false negative respectively, and $A, B$ denote two binary masks. As with the challenge, all metrics are reported at 128 confidence threshold.

\begin{figure*}[!h]
\centering
\includegraphics[width=0.9\linewidth]{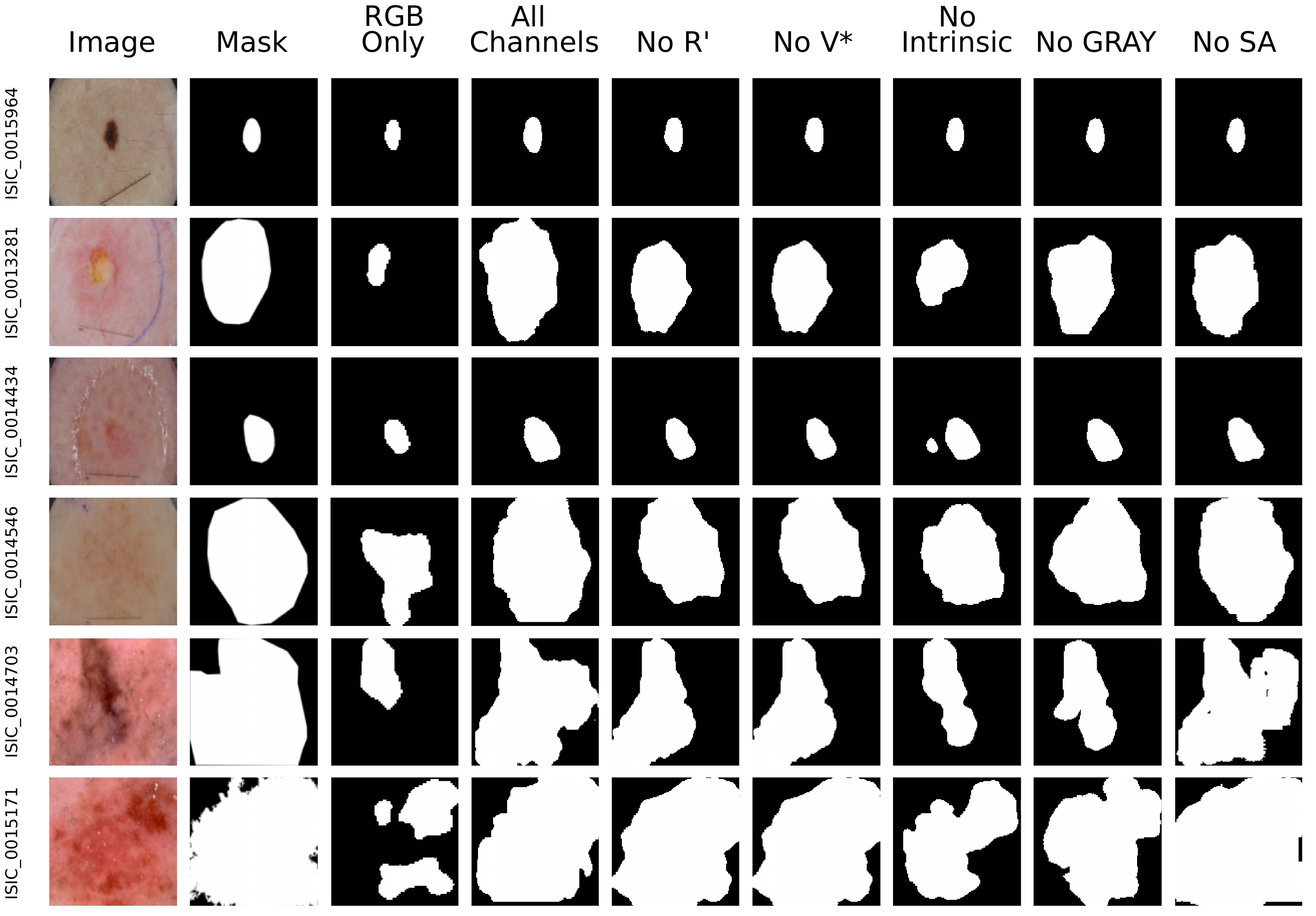}
\caption{Qualitative results for the segmentation performance of all the methods on 6 images from the ISIC 2017 test set. Incorporating the two color bands and illumination-based transformations improves the segmentation consistently, and the performance drop is the most significant when Intrinsic is not used.}
\label{fig:results}
\end{figure*}

\section{Results and Discussion}    \label{sec:Results}

Figure~\ref{fig:TransformedImagesGrid} shows the normalized red channel ($R'$), the complement of the Value channel ($V^*$), the intrinsic image (Intrinsic) using the approach by Finlayson et al.~\cite{finlayson2004intrinsic}, the grayscale converted image (GRAY) and the shading-attenuated image (SA) using the approach by Madooei et al.~\cite{madooei2012automated} for 5 dermoscopic images (with their ISIC image IDs) and their corresponding ground truth segmentation masks from the ISIC 2017 training set. We notice that the presence of artifacts such as markers (second row) and rulers (third and fourth rows) lead to poor results, particularly for shading-attenuated images. While the shading-attenuation results are acceptable for some images, a large number of images yield poor results, such as the last row in Figure~\ref{fig:TransformedImagesGrid}. 

Table~\ref{tab:results} shows the quantitative results for the 600 test images evaluated using the seven trained segmentation networks. We observe that `All Channels' outperforms `RGB Only' in all metrics except specificity, where the difference is quite small. Using all the transformations yields an improvement of 12.02\% and 7.76\% over the baseline (`RGB Only') in the mean Jaccard index and the mean Dice similarity coefficient metrics respectively. We also note that we are within 1\% of the Jaccard index of the top 3 entries on the challenge leaderboard~\cite{leaderboard}, without using any additional external data~\cite{isic3_bi2017automatic}, post-processing, or an ensemble of models~\cite{isic1_yuan2017automatic,isic2_berseth2017isic} and without optimizing the network architecture or any other hyperparameters.

\begin{table*}[!h]
\centering
\caption{Quantitative results for the two methods on the 390 test images from the DermoFit dataset and 200 images from the PH2 dataset (mean $\pm$ standard error).}
\label{tab:results_V2}
\resizebox{\textwidth}{!}{%
{\renewcommand{\arraystretch}{1.5}
\begin{tabular}{ccccccc}
\hline
\textbf{Dataset}                & \textbf{Method}         & \textbf{Accuracy}        & \textbf{Dice Coefficient} & \textbf{Jaccard Index}   & \textbf{Sensitivity} & \textbf{Specificity} \\ \hline
\multirow{2}{*}{DermoFit} & RGB Only & $0.9024 \pm 0.0038$ & $0.8437 \pm 0.0053$ & $0.7418 \pm 0.0069$ & $0.8080 \pm 0.0078$ & $\textbf{0.9534} \pm \textbf{0.0030}$\\  
 & No SA & $\textbf{0.9124} \pm \textbf{0.0030}$ & $\textbf{0.8674} \pm \textbf{0.0042}$ & $\textbf{0.7737} \pm \textbf{0.0055}$ & $\textbf{0.8721} \pm \textbf{0.0053}$ & $0.9347 \pm 0.0032$\\
\hline
\multirow{2}{*}{PH2} & RGB Only & $0.8546 \pm 0.0133$ & $0.7989 \pm 0.0128$ & $0.6944 \pm 0.0140$ & $0.8032 \pm 0.0166$ & $0.9543 \pm 0.0031$\\  
 & No SA & $\textbf{0.8926} \pm \textbf{0.0091}$ & $\textbf{0.8537} \pm \textbf{0.0071}$ & $\textbf{0.7559} \pm \textbf{0.0091}$ & $\textbf{0.8442} \pm \textbf{0.0110}$ & $\textbf{0.9607} \pm \textbf{0.0022}$\\
\hline
\end{tabular}%
}
}
\end{table*}

To further capture the improvement in the segmentation performance, we plot the kernel density estimates of the metrics for all the methods (Figure~\ref{fig:plots}). We use the Epanechnikov kernel to estimate their probability density functions, and the plots have been clipped to the range of the values of the respective metrics. The plots show higher peaks (indicating higher densities) at larger values of all the metrics for the proposed method(s).

Figure~\ref{fig:results} shows six samples from the test dataset and their corresponding ground truth segmentation masks, along with the prediction outputs from the seven models. The samples have been chosen to cover almost all possible variations in the images, such as the size of the lesion, the contrast between the lesion and the surrounding skin, and the presence of artifacts (ruler, gel bubble, etc.). We note that apart from the improved segmentation performance, incorporating the proposed transformations into the input to the model also considerably improves the false positive and the false negative labels.

\begin{figure}[!h]
\centering
\includegraphics[width=0.48\textwidth]{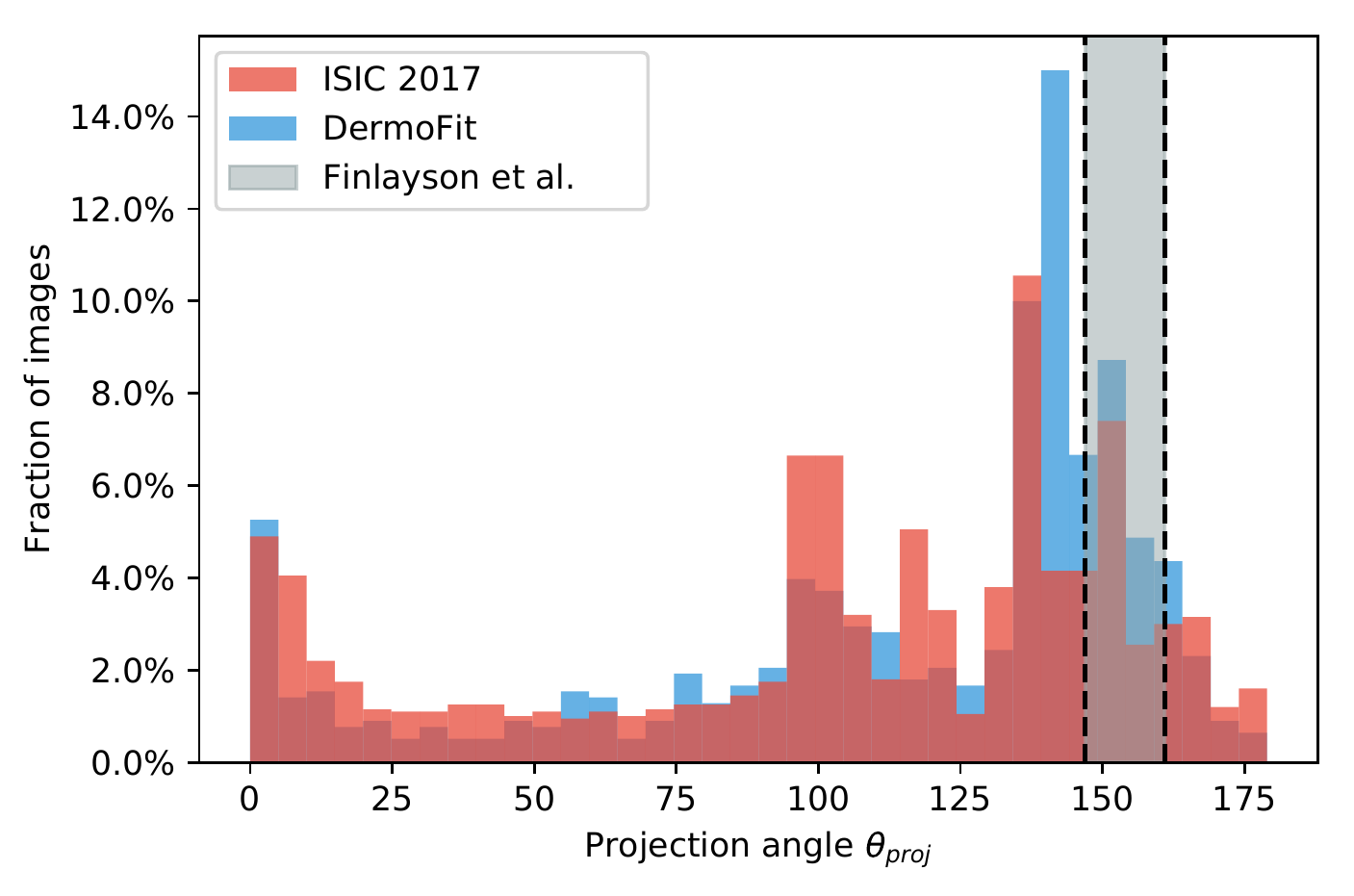}
\caption{Histogram of projection angles for the training images from the ISIC 2017 and the DermoFit datasets. The projection angles for these images are spread across the entire range, whereas it is restricted to a small range for Finlayson et al.~\cite{finlayson2004intrinsic}.}
\label{fig:hist}
\end{figure}

Next, we analyze the contribution of each of the color theory and illumination-based transformations towards improving the segmentation performance. From Table~\ref{tab:results}, we can see that dropping the normalized red channel ($R'$), the complement of the Value channel ($V^*$), and the shading-attenuated image (SA) have the least impact on the Dice coefficients. Of these, the first two can possibly be explained by the fact that these are relatively simpler transformations as compared to the other three, and are therefore easier for the network to learn. As for the SA component, as already noted previously and shown in the SA column in Figure~\ref{fig:TransformedImagesGrid}, a large number of images yield very poor results. Since we use JPEG compressed images, most of the high frequencies (in the Fourier domain representation) are discarded during JPEG compression, which leads to the entropy minimization step producing sub-optimal projection angles. We confirm this by plotting the projection angles calculated for the 2000 and 780 images in the ISIC 2017 and the DermoFit training sets (Figure~\ref{fig:hist}). We observe that the projection angles are spread across the entire range, which is in contrast to Finlayson et al.~\cite{finlayson2004intrinsic} where the minimum entropy angles are between \ang{147} and \ang{161} for their HP912 camera. As such, we do not expect the SA images to provide a considerable improvement when used in a segmentation model, which is consistent with the quantitative results on the ISIC 2017 test set.

On the other hand, we observe that the intrinsic image (Intrinsic) and the grayscale converted image (GRAY) are crucial to the segmentation performance improvement. Since these transformations rely on the log-chromaticity and the optical density space representations respectively, and therefore are not so easily learned by a deep semantic segmentation model. The dip in performance is the most when the Intrinsic image is dropped, indicating that it is the most important illumination-based transformation for improving the segmentation. Figure~\ref{fig:results} shows that `No Intrinsic' also results in higher false positives and false negatives (most clearly visible in the second and the third rows).

Finally, for the DermoFit dataset, we train two models: `RGB Only' and `No SA'. As discussed, SA images 
do not contribute much to improving the segmentation performance (as shown for the ISIC 2017 dataset, Table~\ref{tab:results}), while also being computationally intensive (Algorithm~\ref{alg:shadow}). As such, we use `RGB Only' as the baseline to evaluate the performance of `No SA'. 
As for the PH2 dataset, given the small number of images, we use the entire PH2 dataset as a test set for the two models trained on the DermoFit dataset to evaluate the generalizability of the trained models. 

Table~\ref{tab:results_V2} shows the quantitative results for evaluating these two trained models on the DermoFit test set and the entire PH2 dataset. We observe that `No SA' improves the mean Jaccard index for the DermoFit and the PH2 datasets by 4.30\% and 8.86\% respectively over the `RGB Only' baseline.

\section{Conclusion}    \label{sec:Conclusion}

Motivated by the potential value of leveraging information about the physics of skin illumination and imaging in a data hungry deep learning setting, in this work, we proposed a novel semantic segmentation framework for skin lesion images by augmenting the RGB dermoscopic images with additional color bands and intrinsic, grayscale, and shading-attenuated images. We demonstrated the efficacy of the proposed approach by evaluating on three datasets: the ISIC ISBI 2017 Challenge dataset, the DermoFit Image Library, and the PH2 database and observed a considerable performance improvement over the baseline method. We also performed ablation studies to ascertain the contribution of each of the transformations on the segmentation performance improvement. We hypothesize that, despite being useful for improving prediction accuracy, deep learning does not happen to stumble upon these illumination-based channels given the large search space, the fixed architecture, and the local gradient-descent optimizer. 
Future work may explore architectures, losses, or training strategies that ensure such illumination information are encoded.


{\small
\bibliographystyle{ieee_fullname}
\bibliography{egbib}
}

\end{document}